\def\year{2022}\relax
\documentclass[letterpaper]{article} 
\usepackage{aaai22}  
\usepackage{times}  
\usepackage{helvet}  
\usepackage{courier}  
\usepackage[hyphens]{url}  
\usepackage{graphicx} 
\usepackage{natbib}  
\usepackage{caption} 
\DeclareCaptionStyle{ruled}{labelfont=normalfont,labelsep=colon,strut=off} 
\usepackage{url}
\usepackage{subfigure}
\usepackage{multicol}
\usepackage{multirow}
\usepackage{mathtools}
\usepackage{color}
\usepackage{diagbox}
\usepackage{bbm}
\usepackage{blindtext}
\usepackage{amsmath,amsfonts,amssymb}
\usepackage{bm}
\usepackage{wrapfig}
\usepackage{wrapfig,lipsum,booktabs}
\usepackage{algorithm}
\usepackage{algpseudocode}

\DeclareMathOperator*{\onedcnn}{1D-CNN}

\DeclareMathOperator*{\meanpool}{Pool}
\DeclareMathOperator*{\pool}{Pool}
\DeclareMathOperator*{\mlp}{MLP}
\DeclareMathOperator*{\mysigmoid}{Sigmoid}

\DeclareMathOperator*{\layernorm}{LayerNorm}
\DeclareMathOperator*{\tsalign}{TS-Align}
\DeclareMathOperator*{\tsintegrate}{TS-Integrate}
\DeclareMathOperator*{\multipleSelAttn}{Mul-Sel-Attn}
\DeclareMathOperator*{\SelAttn}{Selective-Attn}

\usepackage{amsmath,amsfonts,bm}

\def\eqref#1{equation~\ref{#1}}

\def\1{\bm{1}}

\def\va{{\bm{a}}}
\def\vb{{\bm{b}}}
\def\vc{{\bm{c}}}

\def\ve{{\bm{e}}}

\def\vg{{\bm{g}}}
\def\vh{{\bm{h}}}

\def\vq{{\bm{q}}}

\def\vs{{\bm{s}}}
\def\vt{{\bm{t}}}
\def\vu{{\bm{u}}}
\def\vv{{\bm{v}}}

\def\vx{{\bm{x}}}

\def\vz{{\bm{z}}}

\def\mC{{\bm{C}}}

\def\mH{{\bm{H}}}

\def\mT{{\bm{T}}}
\def\mU{{\bm{U}}}
\def\mV{{\bm{V}}}
\def\mW{{\bm{W}}}
\def\mX{{\bm{X}}}

\DeclareMathAlphabet{\mathsfit}{\encodingdefault}{\sfdefault}{m}{sl}
\SetMathAlphabet{\mathsfit}{bold}{\encodingdefault}{\sfdefault}{bx}{n}

\def\gB{{\mathcal{B}}}

\def\gD{{\mathcal{D}}}

\def\gL{{\mathcal{L}}}

\def\gT{{\mathcal{T}}}

\def\sV{{\mathbb{V}}}

\def\sZ{{\mathbb{Z}}}

\newcommand{\R}{\mathbb{R}}

\newcommand{\softmax}{\mathrm{softmax}}

\usepackage{newfloat}
\usepackage{listings}
\floatstyle{ruled}
\newfloat{listing}{tb}{lst}{}
\floatname{listing}{Listing}
\title{Hierarchical Relation-Guided Type-Sentence Alignment for Long-Tail Relation Extraction with Distant Supervision}
\author {
    Yang Li,
    Guodong Long,
    Tao Shen, 
    Jing Jiang
}
\affiliations {
    Australian AI Institute, School of CS, FEIT, University of Technology Sydney\\
    \texttt{yang.li-17@student.uts.edu.au, \{guodong.long,tao.shen,jing.jiang\}@uts.edu.au}
}

\usepackage{bibentry}

\begin{document}
\maketitle

\begin{abstract}
Distant supervision uses triple facts in knowledge graphs to label a corpus for relation extraction, leading to wrong labeling and long-tail problems. Some works use the hierarchy of relations for knowledge transfer to long-tail relations. However, a coarse-grained relation often implies only an attribute (e.g., domain or topic) of the distant fact, making it hard to discriminate relations based solely on sentence semantics. One solution is resorting to entity types, but open questions remain about how to fully leverage the information of entity types and how to align multi-granular entity types with sentences. 
In this work, we propose a novel model to enrich distantly-supervised sentences with entity types. It consists of (1) a pairwise type-enriched sentence encoding module injecting both context-free and -related backgrounds to alleviate sentence-level wrong labeling, and (2) a hierarchical type-sentence alignment module enriching a sentence with the triple fact's basic attributes to support long-tail relations. Our model achieves new state-of-the-art results in overall and long-tail performance on benchmarks.
\end{abstract}

\section{Introduction} \label{sec:introduction}

Human-curated knowledge graphs (KGs), play a critical role in many downstream tasks but suffer from the incompleteness \citep{xiong2018knowledge,yao2019kg}.
As a remedy, relation extraction is to distinguish the relation between two entities according to their semantics in text, but a major obstacle is a lack of sufficient labeled corpus. 
Fortunately, distant supervision can be used to annotate a raw text corpus via KGs for relation extraction, a.k.a. distantly supervised relation extraction (DSRE). 
This is based on a strong assumption that a sentence containing two entities will express the semantics of their relation in a KG \citep{riedel2010modeling}. 

\begin{figure}[t]
    \centering
    \includegraphics[width=0.43\textwidth]{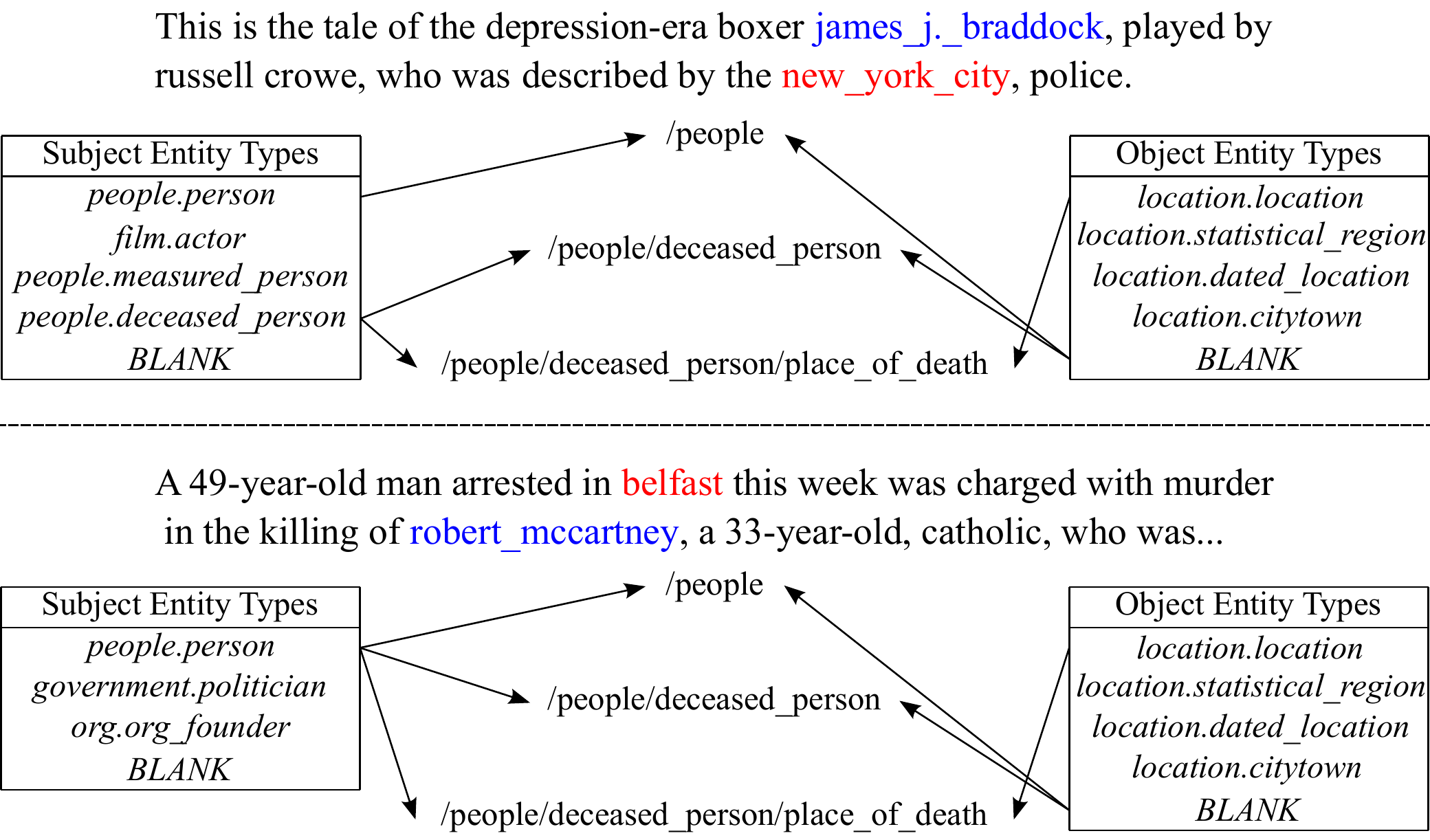}
    \caption{Two sentences with the same long-tail relation. For each sentence, multi-granular relations from top to bottom are pointed by its best pairwise types, which indicates not all pairwise types provide the same contribution.
    \textcolor{blue}{Blue} is subject entity, and \textcolor{red}{red} is object entity.
    The 1st sentence relies on the direct pairwise types due to its relation-irrelevant semantics while the 2nd sentence integrates its relation-relevant semantics and pairwise types to enhance its representation.
    }
    \label{fig:intro}
\end{figure}

The assumption cannot always hold, leading to the wrong labeling problem. For example, both ``\textit{\underline{Jobs} founded \underline{Apple}}'' and ``\textit{\underline{Jobs} ate \underline{Apple}}'' are labeled with ``\textsc{/business/company/founders}'' according to a KG triple fact (\textit{Steven Jobs}, \textsc{/business/company/founders}, \textit{Apple Inc}). A basic technique for this problem is selective attention \citep{zeng2015distant, lin2016neural, ji2017distant} under multi-instance learning framework \citep{riedel2010modeling, hoffmann2011knowledge}. Given a bag of sentences with the same entity pair, it learns to select correct one(s) by an end-to-end attention.
The other major challenge is known as the long-tail problem, caused by domain mismatching during distant supervision. 
That is, many relation labels correspond only to a limited number of training sentences in the corpus \citep{ye2019looking}. 
For example, in a DSRE benchmark, the distant supervision is an encyclopedic KG (i.e., Freebase \citep{bollacker2008freebase}) while the corpus is news articles from the New York Times (NYT), so relations, like ``\textsc{/people/person/religion}'', scarcely appear. 
As illustrated by \citet{li2020improving} and \citet{zhang2019long}, more than $70\%$ of relation labels in NYT can be regarded as \textit{long-tail relations}.

To mitigate the long-tail problem, some works \citep{han2018hierarchical,zhang2019long,li2020improving} resort to the hierarchy of relations for knowledge transfer from data-rich relations to the long-tail ones since the relations have coarse-grained overlap.
They focus on interactive operations between hierarchical relations and intra-bag sentences, including relation-to-sentence attention \citep{han2018hierarchical} as a hierarchical extension of selective attention, and sentence-to-relation attention \citep{li2020improving} enriching sentences with multi-granular relations.
As such, they achieve knowledge transfer by learning to distinguish coarse-grained relations for sentences with sufficient data, which provides a latent constraint for the long-tail relations. 
However, a coarse-grained relation usually denotes the only basic attribute of the distant oracle triple fact in KG, so a sentence scarcely contains its semantics and we can only imply the relation via background information. 
Again, true-labeled ``\textit{Jobs founded Apple}'', does not contain any semantics of its coarse-grained relation ``\textsc{/business/company}'', but we can reason it from the predicate \textit{founded} and type of \textit{Apple}. 
Thus, it is a challenge for a hierarchical DSRE model to correctly imply coarse-grained relations based solely on sentences, not to mention the existence of the wrong labeling problem.

A direct yet promising way to overcome this challenge is to incorporate extra information for entities in a sentence \citep{vashishth2018reside,hu2019improving,chu2020insrl}. 
One popular source is the entity types, i.e., an entity's ``\textsc{isA}'' attributes in KG, which characterizes the entity from multiple perspectives \citep{chen2020improving}.
As Figure \ref{fig:intro} shows, although the 1st sentence's semantics is irrelevant to relation, the pairwise types \textsl{people.deceased\_person} and \textsl{location.location} directly align with the fine grained relation. 
However, existing works \citep{vashishth2018reside,chu2020insrl} ignore this potential of explicit structured types information.

In this work, we aim to improve DSRE by exploiting structured information in the entity types from both pairwise and hierarchical perspectives to alleviate the wrong labeling and the long-tail problems respectively. 
To this end, we first propose a \textit{context-free type-enriched embedding} module to generate word embeddings with pairwise types associated with the entity pair in a bag. As mentioned in Figure \ref{fig:intro}, even without the corresponding semantic support, pairwise types can provide direct attributes of entities to align with the relation.
Besides, we develop a \textit{context-related type-sentence alignment} module to generate robust sentence representation with pairwise types. Since entities have specific characteristics in certain semantics, we leverage semantics to select proper pairwise types and then enrich sentence representation, as the 2nd sentence in Figure \ref{fig:intro} shows.
Such an alignment is enhanced by a guidance from the relation to auto-seek for associations between pairwise types and sentences. 

At the meantime, hierarchical information has been proven crucial in knowledge transfer for long-tail relations \citep{han2018hierarchical,zhang2019long,li2020improving}.
Thereby, we naturally extend the base alignment module into a hierarchy by proposing a \textit{hierarchical type-sentence alignment} module. 
An intuitive example in Figure \ref{fig:intro} shows that different grained relations are pointed by various granular pairwise types. 
This indicates that these pairwise types contain hierarchical semantics, which makes it feasible to extend base alignment into hierarchy.
Thus, the strong association between pairwise types and coarse-grained relations can improve knowledge transfer for long-tail relations. 

We conduct extensive experiments on two popular benchmarks, NYT-520k and NYT-570k, showing that our model achieves new state-of-the-art overall and long-tail performance. Further analyses reveal insights into our model.

\begin{figure*}[t]
    \centering
    \includegraphics[width=0.85\textwidth]{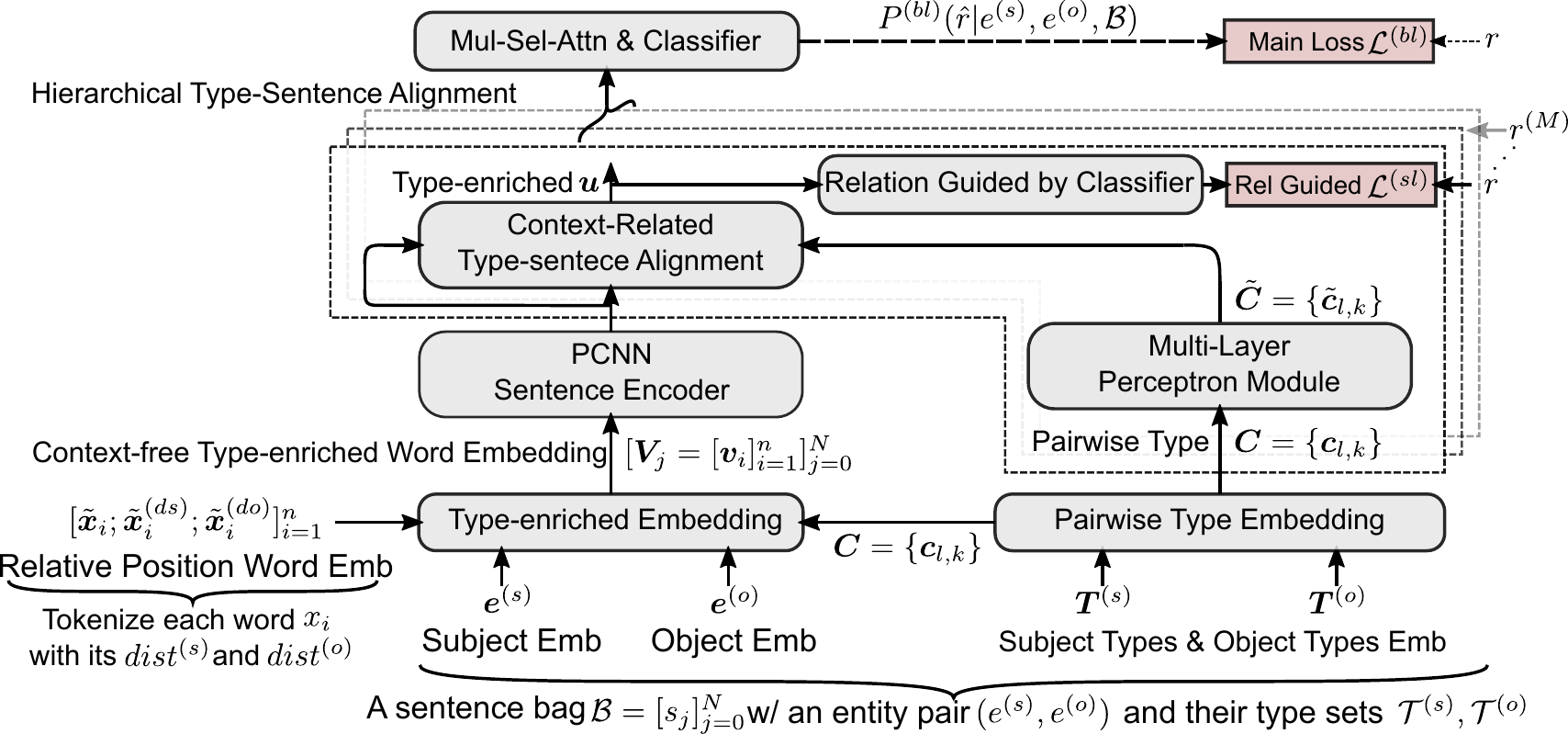}
    \caption{\small Our proposed model, called \textbf{Hi}erarchical \textbf{R}elation-guided Type-Sentence \textbf{A}lignment \textbf{M}odel (HiRAM), for DSRE.
    }
    \label{fig:model}
\end{figure*}

\section{Approach}
In this section, we elaborate on our novel neural network, as illustrated in Figure \ref{fig:model}. 

\paragraph{Task Definition.}
Given a bag of sentences $\gB = \{s_1, \ldots, s_N\}$ containing a pair of subject $e^{(s)}$ and object $e^{(o)}$ entities, the distant supervision \citep{mintz2009distant} assigns the sentence bag with a relation label $r$ according to KG triple fact. 
The goal of relation extraction is to predict the relation label $\hat{r}$ of an entity pair based on the corresponding sentences bag $\gB$. 
Labels of coarse-grained relations, $[r^{(1)}, \dots, r^{(M)}]$, can be derived from the mention of $r$. For instance, when $r=$ \textsc{/business/company/founders}, $r^{(1)} =$ \textsc{/business/company} and $r^{(2)} =$ \textsc{/business}.

\subsection{Context-Free Type-Enriched Word Emb} \label{sec:cf_type_enrich}

Following most previous DSRE works, we first tokenize each sentence $s_j\in\gB$ and employ a word2vec method \citep{mikolov2013distributed} to derive a sequence of word embeddings by looking up a learnable matrix $\mW^{(emb)}\in\R^{d_e\times |\sV|}$, i.e., $\tilde\mX^j = [\tilde\vx^j_1, \dots, \tilde\vx^j_n]\in\R^{d_e}$, where $\sV$ denotes word vocabulary. $j$ denotes the index of a sentence in the bag and $n$ denotes the sentence length. 
In the sequel, we omit $j$ if no confusion is caused. Then, as a common practice in DSRE \citep{zeng2014relation}, a word's relative distances to both the subject and object entities (a.k.a relative positions) also play significant roles. The distances are first denoted as two integers ($dist^{(s)}$ and $dist^{(o)}\in\sZ$) and then embedded into two learnable vectors ($\tilde\vx^{(ds)}_i$ and $\tilde\vx^{(do)}_i\in\R^{d_p}$). 
Therefore, the updated sequence of word embeddings is $\mX^j = [\vx_1, \dots, \vx_n]$, where $\vx_i=[\tilde\vx_i; \tilde\vx^{(ds)}_i; \tilde\vx^{(do)}_i]\in\R^{d_w}$, $[;]$ denotes vector concatenation, and $d_w\coloneqq d_e+2d_p$.

Previous works \citep{li2020self,li2020improving} also found that explicitly enriching each word with both entity embeddings (i.e., $\ve^{(s)}$ and $\ve^{(o)}$) in a context-free manner is also important to DSRE's success.
However, many entities scarcely appear in the raw corpus and have faced polysemy problem (e.g., \textit{Apple} could be a fruit or a company). Thus, model is hard to learn the characteristic of entities and sentence semantics.

Therefore, we leverage entity types to characterize entities' attributes.
That is, given an entity $e$, its types are defined as a set of type mentions, i.e., $\gT = \{t_1, t_2, \dots\}$.
However, previous works \cite{chu2020insrl} directly concatenate the entity types of both $e^{(s)}$ and $e^{(o)}$, completely regardless of potentials of explicit structured information of types.
As demonstrated by \citet{krompass2015type}, a relation in KG is usually constrained by the entity types of $e^{(s)}$ and $e^{(o)}$ simultaneously (i.e., pairwise types), instead of their individuals. 
We thereby propose a pairwise type embedding module to enrich the word embedding $\mX$ also in a context-free manner. 

\paragraph{Type and Pairwise Type Embedding. } First, given an entity type set $\gT= \{t_1, t_2, \dots\}$ (either $\gT^{(s)}$ for subject or $\gT^{(o)}$ for object), we tokenize each type mention $t_j$ into a sequence of words, then embed the words by looking up $\mW^{(emb)}$, and lastly derive the type embedding $\vt_j$ by applying a mean-pooling to the word embeddings of the mention. The embedding of the entire type is
\begin{align}
    \mT = [\vt_1, \vt_2, \dots]\in\R^{|\gT|\times d_e}. \label{eq:tp_start}
\end{align}
As such, we subsequently define the embedding of the pairwise type by considering a combination of every subject $\forall t^{(s)}_l\in\gT^{(s)}$ and object type $\forall t^{(o)}_k\in\gT^{(o)}$. 
Instead of sole semantics via a vector concatenation, we take into account the structured information in each type pair by leveraging a translational scheme \citep{bordes2013translating}. Hence, we represent each type pair ($t^{(s)}_l, t^{(o)}_k$) as
\begin{align}
    \vc_{l,k} &= [\tilde\vc^{(sem)}_{l,k};\tilde\vc^{(str)}_{l,k}]\in\R^{4d_e}, \\
    \notag &~\text{where,}~\tilde\vc^{(sem)}_{l,k} = \vt^{(s)}_l \odot \mW^{(sem)}\vt^{(o)}_k,\\
    \notag&~~~~~~\text{and}~\tilde\vc^{(str)}_{l,k} = \vt^{(o)}_k - \vt^{(s)}_l.
\end{align}
Here, ``$\odot$'' denotes Hadamard product, and $\mW^{(sem)}$ denotes a learnable projection. $\tilde\vc^{(sem)}_{l,k}$ aims to capture the semantic relation in the pair \citep{nickel2011three} since not all types combinations are valid in the whole dataset.
$\tilde\vc^{(str)}_{l,k}$ aims to measure its structured relation. Lastly, we denote all the embeddings of pairwise types as
\begin{align}
    \mC = \{\vc_{l,k}\}_{\forall l \in [1, |\gT^{(s)}|], \forall k \in [1, |\gT^{(o)}|]}, \label{eq:tp_end}
\end{align}
where $\mC\in\R^{4d_e\times m}$ and $m=|\gT^{(s)}|\cdot |\gT^{(o)}|$. 


\paragraph{Type-Enriched Word Embedding. } However, an open question still remains about how to operate on variable-length embeddings of pairwise types, $\mC$, to enrich each word embedding, $\vx_j\in\mX$, in a context-free manner. Inspired by self-attentive sentence encoding \citep{lin2016neural}, we present a bag-level type-attentive module, which compresses $\mC$ into a single vector representation to facilitate type-enriching. 
Intuitively, such self-attentive module is focused on the prior knowledge of the type pair in the corpus. 
Formally, we first generate a global query \citep{lin2016neural} also with structured information of both entities and types, i.e., 
\begin{align}
    \tilde \vq^{(f)} \!\! = \![\ve^{(o)};\! \meanpool(\mT^{(o)})] \!-\! [\ve^{(s)};\! \meanpool(\mT^{(s)})], \label{eq:global_query}
\end{align}
followed by a standard Bilinear-based attention, 
\begin{align}
    \vq^{(f)} \!=\! \mC \cdot \softmax(\mC^T \mW^{(sa)} \vq^{(f)})\!\in\!\R^{4d_e},
\end{align}
where ``$\cdot$'' denotes matrix multiplication and $\mW^{(sa)}$ is a learnable weight matrix of such self-attentive module. Lastly, we use a gate as in \citep{li2020improving} to derive the context-free type-enriched word embedding, i.e., 
\begin{align}
    &\vg^{(gf)}_i = \mysigmoid(\mlp([\vx_i; \vq^{(f)}];\theta^{(gf1)})), \\
    &\vx^{(gf)}_i = \mlp([\vx_i; \vq^{(f)}];\theta^{(gf2)}), \\
    &\vv_i = \vg^{(gf)}_i \odot \vx_i + (\bm 1-\vg^{(gf)}_i)\odot \vx^{(gf)}_i,
\end{align}
where $\mlp$ denotes a multi-layer perceptron (MLP) module. 
Hence, word embeddings for $s$ are updated to $\mV=[\vv_1, \dots, \vv_n]\in\R^{d_w\times n}$.

\subsection{Context-Related Type-Sentence Alignment} \label{sec:cr_tsa}

\paragraph{Sentence Encoding. } 
In DSRE, piecewise convolutional neural network (PCNN) \citep{zeng2015distant} is used to obtain sentence embedding. 
1D-CNN \citep{kim2014convolutional} is first invoked over $\mV$ for contextualized representations. Then a piecewise max-pooling performs over the output sequence to obtain sentence-level embedding with highlighted entity positions:
\begin{align}
    &\notag \mH = [\vh_1, \ldots, \vh_n] = \onedcnn(\mV; \theta^{(cnn)}), \\
    &\notag \vs \!=\! \tanh([\pool(\mH^{(1)});\!\pool(\mH^{(2)});\!\pool(\mH^{(3)})]),
\end{align}
where 
$\mH^{(1)}$, $\mH^{(2)}$ and $\mH^{(3)}$ are three consecutive parts of $\mH$ by dividing $\mH$ w.r.t. the indices of subject $e^{(s)}$ and object $e^{(o)}$ entities. Consequently, $\vs \in \R^{d_h}$ is the resulting sentence-level embedding.

\paragraph{Type-Sentence Alignment.} Consider that types are not comprehensive enough to align with multi-granular relations, we leverage semantic context to select valid pairwise types for generating robust sentence representation.
Hence, we first calculate alignment scores between a sentence $\vs\in\R^{d_h}$ and the embeddings of pairwise types $\mC\in\R^{4d_e\times m}$ by using a simple Bilinear layer, i.e.,  
\begin{align}
    \tilde \mC &= \mlp(\mC; \theta^{(p)}) \in\R^{d_h\times m}, \label{eq:align_start} \\
    \va &= \softmax(\tilde\mC^T \mW^{(al)}\vs)\in\R^m. \label{eq:align_end}
\end{align}
Then, 
we enrich the sentence embedding with the aligned type pairs via another gating mechanism:
\begin{align}
    & \vz = \tilde\mC\cdot\va \label{eq:integrate_start}\\
    &\vg = \mysigmoid(\mlp([\vs; \vz];  \theta^{(g)})), \\
    &\tilde\vu = \vg \odot \vs + (1-\vg) \odot \vz.
\end{align}
Lastly, following previous success \citep{li2020improving,devlin2018bert}, we leverage a residual connection \citep{he2016deep} with layer normalization \citep{ba2016layer} to derive the final context-related type-enriched sentence embedding, i.e.,
\begin{align}
    \vu = \layernorm(\vs + \tilde\vu; \theta^{(lm)}). \label{eq:integrate_end}
\end{align}


\paragraph{Relation-Guided Alignment at the Sentence Level. }
Due to the severe wrong labeling problem at the sentence level, previous DSRE works usually skip over sentence-level relation supervisions.
Fortunately, empowered by the proposed context-free type enrichment and context-related type-sentence alignment, we can utilize the sentence-level relation label even if the relation label is wrong. 
The reason for this is that, a sentence has already been equipped with structured background to support sentence-level relation even if the sentence semantics cannot deliver the relation.
We applied an MLP-based neural classifier to the type-enriched sentence embedding, $\vu$, to determine the relation at the sentence level, i.e., 
\begin{align}
  P^{(sl)}(\hat r |\vu) = \softmax(\mlp(\vu; \theta^{(sl)})),
\end{align}
where, $P^{(sl)}(\hat r |\vu)$ is a categorical distribution over all possible relations. 
Hence, the training objective is to minimize the cross-entropy loss,
\begin{align}
    \gL^{(sl)} = - \sum_\gD\sum_\gB 
    \log P^{(sl)}(\hat r=r |\vu), \label{eq:rel_guidance}
\end{align}
where $\gD$ denotes a DSRE dataset consisting of sentence bags $\gB$.
The guidance from the sentence-level relation leads to 
strong type-sentence alignment (as illustrated in \S\ref{sec:overall} and \S\ref{sec:ablation_study}). As a result, the sentence-level wrong labeling problem is alleviated, which thus contributes in the final bag-level classification.  
In contrast, previous works w/ sentence-level relation supervisions \citep{li2002learning} suffer from the confirmation bias problem \citep{chen2019improving} caused by the sentence-level wrong labeling.

\begin{table*}[t]\small
	\centering
    \setlength{\tabcolsep}{1.8pt}
	\begin{tabular}{lcccccccccccc|c}
		\hline
		\multirow{2}{*}{ \textbf{P@N (\%)} }&\multicolumn{4}{c}{\bf One}&\multicolumn{4}{c}{\bf Two}&\multicolumn{4}{c|}{\bf All}&\multirow{2}{*}{\textbf{AUC}}\\
		\cline{2-13} 
		 &100&200&300&Mean&100&200&300&Mean&100&200&300&Mean&\\
		\hline
		\multicolumn{13}{l}{\textit{Comparative Approaches} } & \\
		\hline
		CNN+ATT \cite{lin2016neural} &76.2&65.2&60.8&67.4&76.2&65.7&62.1&68.0&76.2&68.6&59.8&68.2&-\\
		PCNN+ATT \cite{lin2016neural} &73.3&69.2&60.8&67.8&77.2&71.6&66.1&71.6&76.2&73.1&67.4&72.2&0.341\\
		CoRA \cite{li2020improving} &78.0&69.0&66.0&71.0&79.0&72.0&66.3&72.4&81.0&74.0&68.3&74.4&0.344\\
		RESIDE \cite{vashishth2018reside}
		&80.0&75.5&69.3&74.9&83.0&73,5&70.6&75.7&84.0&78.5&75.6&79.4&-\\
		InSRL \cite{chu2020insrl} &-&-&-&-&-&-&-&-&-&-&-&-&0.451\\
		\midrule
		\textbf{HiRAM}
		&\textbf{93.0}&\textbf{89.0}&\textbf{83.0}&\textbf{88.3}&\textbf{93.0}&\textbf{88.5}&\textbf{84.0}&\textbf{88.5}&93.0&\textbf{88.5}&86.0&\textbf{89.2}&\textbf{0.484}\\
		\midrule [0.2ex]
		\multicolumn{13}{l}{\textit{Ablations} } & \\
		\hline
		HiRAM w/o Hierarchy in \S \ref{sec:hier}
		&88.0&84.5&83.0&85.2&90.0&86.0&85.0&87.0&90.0&86.5&85.0&87.2&0.450\\
		HiRAM w/o CFTE in \S \ref{sec:cf_type_enrich} 
		&78.0&75.5&74.3&75.9&87.0&76.5&74.0&79.2&87.0&77.5&74.7&79.7&0.425\\
		HiRAM w/o Rel Guidance in Eq. \ref{eq:rel_guidance}
		&89.0&86.0&76.7&83.9&\underline{93.0}&88.0&81.7&87.6&\underline{94.0}&87.0&\underline{86.7}&\underline{89.2}&0.482\\
		HiRAM w/ Type Concat
		&84.0&82.0&75.3&80.4&85.0&81.5&79.7&82.1&89.0&82.5&78.0&83.2&0.462\\
		HiRAM w/ BERT-base
		&86.0&84.5&81.7&84.1&86.0&83.0&79.7&82.9&86.0&82.0&79.3&82.4&\underline{0.529}\\
		\hline
	\end{tabular}
	\vspace{-1mm}
	\caption{\small Model Evaluation and ablation study on NYT-520K. ``P@N'' (top-n precision) denotes precision values for the entity pairs with the top-100, -200 and -300 prediction confidences by randomly keeping one/two/all sentence(s) in each bag. Comprehensive analysis of ``P@N'' and ``AUC'' reflects precision and confidence of correct predictions. 
	``HiRAM w/o Hierarchy'' denotes the use of the base model in \S \ref{sec:cr_tsa}. The abbreviation of ``CFTE'' represents the \textbf{C}ontext-\textbf{F}ree \textbf{T}ype-\textbf{E}nriched Word 
	Embedding in \S \ref{sec:cf_type_enrich}. ``HiRAM w/o Rel Guidance'' denotes removing the relation guidance in type-sentence alignment, and ``HiRAM w/ Type Concat'' replaces pairwise types embedding Eq.(\ref{eq:tp_start}-\ref{eq:tp_end}) with its simple concatenation. Finally, ``HiRAM w/ BERT-base'' represents us replacing the embedding layer (in \S \ref{sec:cf_type_enrich}) with the BERT-base model.}
	\label{tab:eval_on_nyt520k}
\end{table*}

\begin{table*}[t]\small
	\centering
    \setlength{\tabcolsep}{1.5pt}
	\begin{tabular}{lcccccccccccc|c}
		\hline
		\multirow{2}{*}{ \textbf{P@N (\%)} }&\multicolumn{4}{c}{\bf One}&\multicolumn{4}{c}{\bf Two}&\multicolumn{4}{c|}{\bf All}&\multirow{2}{*}{\textbf{AUC}}\\
		\cline{2-13} 
		 &100&200&300&Mean&100&200&300&Mean&100&200&300&Mean&\\
		\hline
		\multicolumn{13}{l}{\textit{Comparative Approaches} } & \\
		\hline
		PCNN+HATT \cite{han2018hierarchical} &84.0&76.0&69.7&76.6&85.0&76.0&72.7&77.9&88.0&79.5&75.3&80.9&0.42\\
		PCNN+BAG-ATT \cite{ye2019distant}  &86.8&77.6&73.9&79.4&91.2&79.2&75.4&81.9&91.8&84.0&78.7&84.8&0.42\\
		SeG \cite{li2020self} 
		&94.0&89.0&85.0&89.3&91.0&89.0&87.0&89.0&93.0&90.0&86.0&89.3&0.51\\
		CoRA \cite{li2020improving} 
		&94.0&90.5&82.0&88.8&98.0&91.0&86.3&91.8&\textbf{98.0}&92.5&88.3&92.9&0.53\\
		\midrule
		\textbf{HiRAM}
		&\textbf{96.0}&\textbf{91.5}&\textbf{85.7}&\textbf{91.1}&\textbf{98.0}&\textbf{94.5}&\textbf{89.3}&\textbf{93.9}&\textbf{98.0}&\textbf{95.0}&\textbf{92.3}&\textbf{95.8}&\textbf{0.580}\\
		\bottomrule 
	\end{tabular}
	\vspace{-1mm}
	\caption{\small Model Evaluation on NYT-570K, published by PCNN+HATT \cite{han2018hierarchical}}.
	\label{tab:eval_on_nyt570k}
\end{table*}

\subsection{Hierarchical Type-Sentence Alignment} \label{sec:hier}

Inspired by former works \citep{han2018hierarchical,zhang2019long,li2020improving} for handling long-tail relations, we also extend our basic model into hierarchy. 
However, the basic attributes contained by coarse-grained relation are irrelevant to the semantics in sentences. 
Thus, instead of direct operating on the hierarchy of relations (i.e., from fine-grained $r$ to coarse-grained $[r^{(1)} \dots r^{(M)}]$ relations), we leverage coarse-grained entity types describing the domain/type properties of the entities in the triple facts to enrich each sentence via the guidance from coarse-grained relation because such multi-granular pairwise types are on par with the relation hierarchy.

Formally, we adapt the relation-guided type-sentence alignment (\S\ref{sec:cr_tsa}) into hierarchy, which shares a high-level inspiration with multi-head attention \citep{vaswani2017attention}. First, we reuse the architecture from Eq.(\ref{eq:align_start}-\ref{eq:integrate_end}) by defining
\begin{align}
    \notag \va^{(l)}, \tilde \mC^{(l)} &= \tsalign\nolimits^{(l)}(\vs, \mC),~\forall l\in[1, M], \\
    \vu^{(l)} &= \tsintegrate\nolimits^{(l)}(\va^{(l)}, \tilde \mC^{(l)}, \vs), \label{eq:for_case_study} 
\end{align}
where $\tsalign()$ denotes Eq.(\ref{eq:align_start}-\ref{eq:align_end}) to obtain type-sentence alignment $\va^{(l)}$ and $\tsintegrate()$ denotes Eq.(\ref{eq:integrate_start}-\ref{eq:integrate_end}) to generate enriched sentence representation $\vu^{(l)}$ at level $l$. Note that, these modules are parameter-untied from each other. Then, we update the sentence-level relation-guided loss in Eq.(\ref{eq:rel_guidance}) to its hierarchical version, i.e., 
\begin{align}
    \gL^{(sl)} = -\!\!\!\!\!\!\!\!\!\!\sum_{\gD,\gB,l\in[1,M]} \!\!\!\!\!\!\!\!\log P^{(sl)}(\hat r^{(l)}\!\!=\!\!r^{(l)}|\vu^{(l)}) \label{eq:hier_rel_guidance}
\end{align}
Again, learnable parameters of the sentence-level classifiers across $l$ are also untied. 
Lastly, we obtain the hierarchical type-enriched representation, i.e., 
\begin{equation} \label{eq:hier_u}
    \vu^{(h)} = [\vu; \vu^{(1)}; \dots; \vu^{(M)}]\in\R^{(1+M) d_h}.
\end{equation}
Different to previous works \citep{han2018hierarchical,zhang2019long,li2020improving} focusing on hierarchical relation embeddings, our work explores the constraints by pairwise types for relations to mitigate sentence-level wrong labeling and uses the hierarchy of entity types on par with that of the relation to improve long-tail performance.

\subsection{Relation Classification and Objectives}

Lastly, we put the sentences back into the bag and derive bag-level embedding for the final relation classification. Hence, for a bag $\gB=[s_1, ...s_N]$, we can obtain sentence embeddings of all the sentences $\mU^{(h)} = [\vu^{(h)}_1, \dots, \vu^{(h)}_N]$, where $\vu^{(h)}_j$ is hierarchical type-enriched sentence encoding derived from Eq.(\ref{eq:hier_u}).
To preserve the hierarchical information learned in $\vu^{(h)}_j$, we proposed to apply multiple selective modules to its different parts, i.e.,
\begin{align}
    \notag &\vb = \multipleSelAttn(\mU^{(h)}) = [\vb^{(0)}; \vb^{(1)}; \dots; \vb^{(M)}], \\
    \notag &\vb^{(0)} = \SelAttn([\vu_1; \!\dots, \!\vu_N]), \\
    \notag &\vb^{(l)} \!=\! \SelAttn([\vu^{(l)}_1; \dots, \vu^{(l)}_N]),~\forall l \in [1, \!M].  
\end{align}
where, $\SelAttn()$ represents the selective attention among the sentences in each granular relation, and $\multipleSelAttn()$ represents the selective attention among the multi-granular bag representations.
$\vb^{(0)}$ denotes the fine grained bag representation and $\vb^{(l)}$ denotes the coarse grained bag representations. 
Lastly, we use an MLP-based classifier upon $\vb$ to derive a bag-level categorical distribution, i.e., 
\begin{align}
    P^{(bl)}(\hat r| e^{(s)}, e^{(o)}, \gB).
\end{align}
Meanwhile, the corresponding training loss is 
\begin{align}
    \gL^{(bl)} = - \sum_\gD P^{(bl)}(\hat r = r| e^{(s)}, e^{(o)}, \gB). \label{eq:bag_loss}
\end{align}
Therefore, the final training objective is to minimize a linear combination of both sentence-level in Eq.(\ref{eq:rel_guidance}) and bag-level (in Eq.(\ref{eq:bag_loss})) losses, i.e., 
\begin{align}
    \gL = \gL^{(bl)} + \beta\gL^{(sl)}. 
\end{align}

\begin{table}[b] \small
    \centering
    \begin{tabular}{ccccc}
        \toprule
        \# Dataset & \# Sentences & \# Entity pairs & \# Relational fact \\ 
        \midrule
        NYT-520K & 522,611 & 281,270 & 18,252 \\ 
        NYT-570K & 570,088 & 293,003 & 19,429 \\
        \bottomrule
    \end{tabular}
    \caption{\small Statistics of NYT training datasets.}
    \label{tab:nyt_statistic}
\end{table}


\begin{table*}[t]\small
    \centering
    \begin{tabular}{lcccccc}
        \toprule
        \multicolumn{1}{l}{\bf \# Training Instance}&\multicolumn{3}{c}{\bf \textless100}&\multicolumn{3}{c}{\bf \textless200}\\ \midrule
        \multicolumn{1}{l}{\textbf{Hits@K}  (Macro)}&\textbf{10}&\textbf{15}&\textbf{20}&\textbf{10}&\textbf{15}&\textbf{20}\\
        \midrule
         PCNN+ATT \cite{lin2016neural} & \textless5.0&7.4&40.7&17.2&24.2&51.5 \\
         PCNN+HATT$^*$ \cite{han2018hierarchical} & 29.6&51.9&61.1&41.4&60.6&68.2 \\
         PCNN+KATT$^*$ \cite{zhang2019long} & 35.3&62.4&65.1&43.2&61.3&69.2 \\ 
         CoRA$^*$ \cite{li2020improving} & 66.6&72.0&87.0&72.7&77.3&89.4 \\
         CoRA \cite{li2020improving} & 66.6&66.6&75.9&71.7&72.7&80.3 \\
        \midrule
         HiRAM & \textbf{72.2}&\textbf{96.3}&\textbf{96.3}&\textbf{77.3}&\textbf{96.9}&\textbf{96.9} \\
        \midrule
         HiRAM w/o Hierarchy in \S \ref{sec:hier} &50.0&88.9&92.6&59.1&90.9&93.9 \\
         HiRAM w/o CFTE in \S \ref{sec:cf_type_enrich}
         &66.6&88.9&92.6&72.7&90.9&93.9\\
         HiRAM w/o Rel Guidance in Eq. \ref{eq:rel_guidance} &55.6&66.7&88.9&63.6&72.7&90.9\\
         HiRAM w/ Type Concat &72.2&77.7&88.9&77.3&81.8&90.9\\
         HiRAM w/ BERT-base
         &55.6&65.1&71.7&63.6&72.7&88.9\\
        \bottomrule
    \end{tabular}
    \vspace{-1.5mm}
    \caption{ \small Hits@K (Macro) tests only on the relations whose number of training instance $<$ 100/200.
    ``Hits@K'' denotes whether a test sentence bag whose gold relation label $r^{(0)}$ falls into top-$K$ relations ranked by their prediction confidences.``Macro'' denotes macro average is applied regarding relation labels. 
    ``$^*$'' denotes the model is trained on NYT-570K.
    }
    \label{tab:hits@k}
\end{table*}

\begin{table*}[] \small
    \centering
    \begin{tabular}{p{150pt}|p{150pt}|p{150pt}}
        \toprule
        \multicolumn{3}{l}{\textbf{Case Sentence 1:} although the regime of president \textbf{bashar\_al-assad} hails from an obscure offshoot of shiism -- the alawites -- syria} \\
        \multicolumn{3}{l}{is nearly three-quarters sunni, with alawites, members of other \textbf{muslim} sects and a considerable number of christians making up the rest.} \\
        \hline
         \multicolumn{1}{c|}{$r^{(2)}$: /people} & \multicolumn{1}{c|}{$r^{(1)}$: /people/person} & \multicolumn{1}{c}{$r^{(0)}$: /people/person/religion}\\
         
        \midrule
        \midrule
         \multicolumn{3}{l}{\textbf{Case Sentence 2:} having so many operating systems makes it expensive to make software \textbf{, said faraz\_hoodbhoy}, the chief executive of } \\
         \multicolumn{3}{l}{camera phones save and share multimedia content.} \\
         \hline
         \multicolumn{1}{c|}{$r^{(2)}$: /business} & \multicolumn{1}{c|}{$r^{(1)}$: /business/company} & \multicolumn{1}{c}{$r^{(0)}$: /business/company/founder}\\
        \bottomrule
    \end{tabular}
    \vspace{-1.5mm}
    \caption{\small Two cases with long-tail relations are mis-classified by previous works whereas HiRAM is competent. Analysis of the attention probability shown in Figure \ref{fig:case_study} proves the effectiveness of context-related type-sentence alignment with relation guidance.}
    \label{tab:case_study}
\end{table*}

\begin{figure*}[t]
    \centering
    \subfigure{\includegraphics[width=0.26\textwidth]{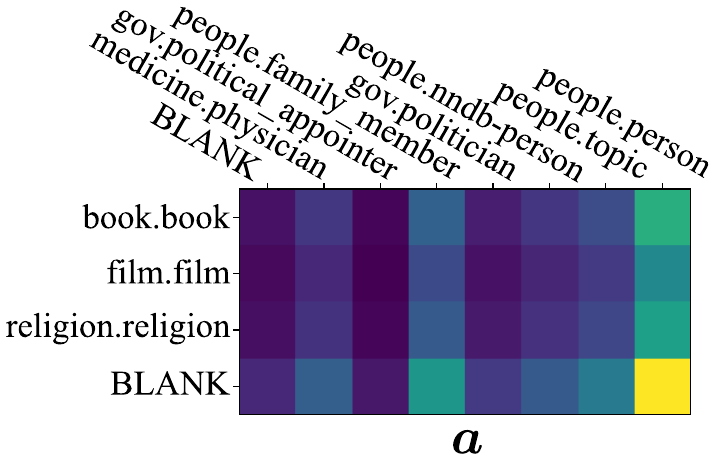}}
    \subfigure{\includegraphics[width=0.26\textwidth]{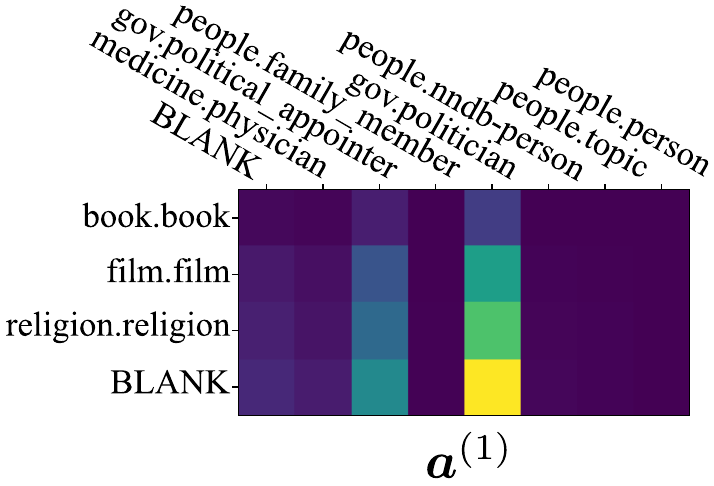} \vspace{-10mm}}
    \subfigure{\includegraphics[width=0.26\textwidth]{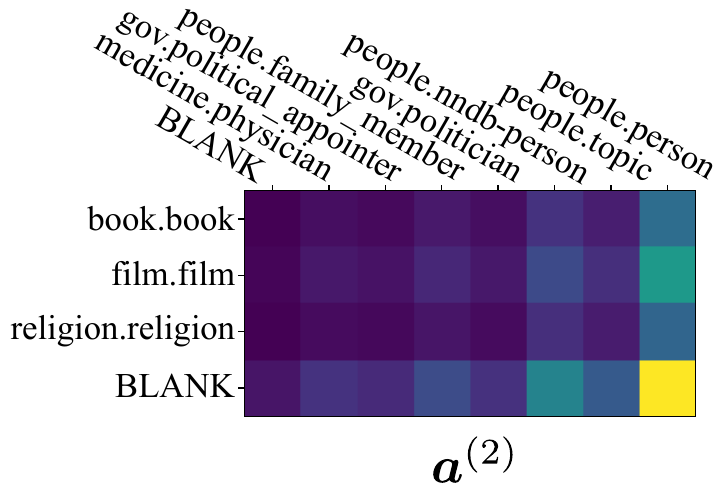} \vspace{-10mm}} \\[-3ex]
    \subfigure{\includegraphics[width=0.26\textwidth]{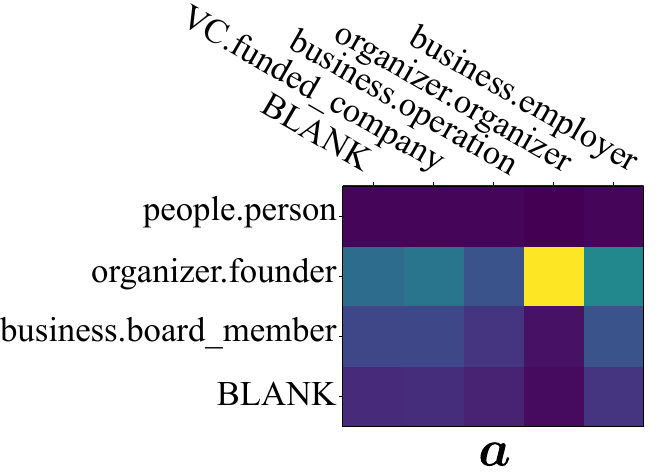} \vspace{-10mm}}
    \subfigure{\includegraphics[width=0.26\textwidth]{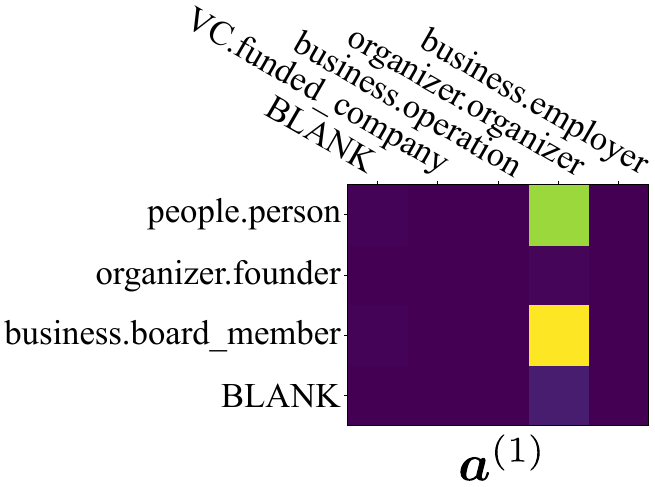}}
    \subfigure{\includegraphics[width=0.26\textwidth]{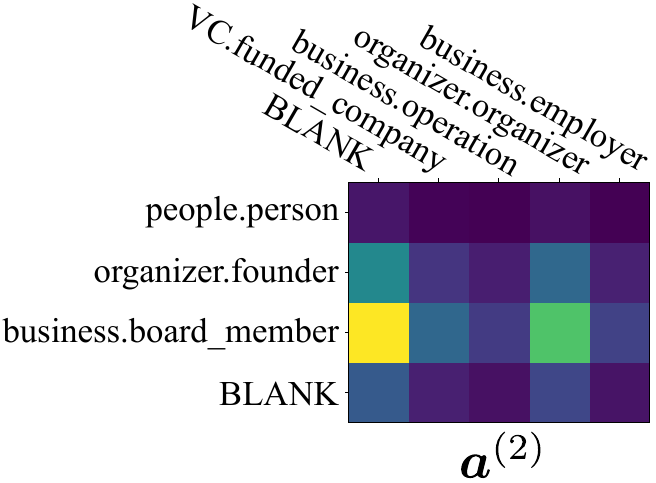}} 
    \vspace{-1.2mm}
    \caption{Each heatmap represents the distribution of type-sentence alignment $\va$ in Eq.(\ref{eq:align_end}) and $\va^{l}$ in Eq.(\ref{eq:for_case_study}). The horizontal axis represents the types of subject entity, and the vertical axis represents the types of object entity. The top row, from left to right, represents three alignment distributions of first case, and the bottom row represents three alignment distributions of second case, as Table \ref{tab:case_study} shows. Notice that ``VC'' is the abbreviation of venture captial.}
    \label{fig:case_study}
\end{figure*}

\section{Experiments}
We evaluate our HiRAM on DSRE benchmarks, New York Times -- NYT \citep{riedel2010modeling}, including NYT-520K and NYT-570K. 

\paragraph{Datasets.} 
NYT datasets have 53 distinct relations, including an \textit{NA} class denoting the unavailable relation between entity pairs. 
As in Table \ref{tab:nyt_statistic}, the difference between NYT-520K and NYT-570K is the number of training sentences, and there is an overlap of 11,416 entity pairs between training and testing in NYT-570K. 
Their common testing set contains 172,448 sentences, with 96,678 entity pairs. 
Compared to NYT-570K, NYT-520K has severer wrong labeling and long-tail problems, and is thus our main test set. 
NYT offers two coarse-grained relations (i.e., $M=2$), and the number of distinct relations from fine to coarse are 53, 36 and 9.

\paragraph{Evaluation Metrics.} Following previous works \citep{lin2016neural,han2018hierarchical,zhang2019long,li2020improving,chu2020insrl}, we use area under precision-recall curve (AUC) and top-N precision (P@N) to measure models' performance with the disturbance of wrong labeling and use Hits@K to measure the performance on long-tail relations. 

\paragraph{Settings.} For both versions of NYT datasets, $d_e$, $d_p$, $d_w$, $d_h$ and $M$ are 50, 5, 60, 690, and 2 respectively. 
The types number of each entity is various but we set an upper limit and pad BLANK as a choice.
We use mini-batch SGD with AdaDelta \cite{zeiler2012adadelta} with $0.1$ learning rate. Batch size is 160 with 15 epochs and 5-th is the best, dropout probability is 0.5, weight decay of L2-reg is $10^{-5}$. We use single Titan XP for computations, except for BERT w/ RTX6000. We use Xavier uniform or BERT-base to initialize our models.

\paragraph{Comparative Approach.} We compare our HiRAM with many strong competitors, including
	\textbf{(1)} \textbf{PCNN+ATT} \citep{lin2016neural} proposes a selective attention to alleviate wrong labeling.
	\textbf{(2)} \textbf{PCNN+HATT} \citep{han2018hierarchical} extends selective attention with hierarchical relations. 
	\textbf{(3)} \textbf{RESIDE} \citep{vashishth2018reside} leverages side KGs' information to improve DSRE.
	\textbf{(4)} \textbf{PCNN+BAG-ATT} \citep{ye2019distant} proposes intra-bag and inter-bag attentions to handle the wrongly labeled sentences.
    \textbf{(5)} \textbf{PCNN+KATT} \citep{zhang2019long} integrates externally pre-trained graph embeddings with relation hierarchies for long-tail relations. 
    \textbf{(6)} \textbf{SeG} \citep{li2020self} focuses on one-sentence bags and proposes entity-aware embedding.
    \textbf{(7)} \textbf{CoRA} \citep{li2020improving} transfers multi-granular relations features into sentences in hierarchies for long-tail relations.
    \textbf{(8)} \textbf{InSRL} \citep{chu2020insrl} integrates sentence, entity description and types together via intact space representation learning. 

\subsection{Overall Performance on Benchmarks} \label{sec:overall}
As shown in Tables \ref{tab:eval_on_nyt520k} and \ref{tab:eval_on_nyt570k}, HiRAM outperforms former baselines on NYT-570K.
Different from CoRA's poor performance on NYT-520K, HiRAM achieves a new state-of-the-art on both popular benchmarks in P@N and AUC.
Compared with InSRL integrating both clean entity types' concatenation and accurate entity descriptions, HiRAM increases the AUC score by nearly 7\%, verifying the capability of pairwise types and hierarchical type-sentence alignment.
 
\subsection{Ablation Study} \label{sec:ablation_study}
We conduct an ablation study on NYT-520K, as shown at the bottom of Table \ref{tab:eval_on_nyt520k}.
Compared to HiRAM, ``HiRAM w/o Hierarchy'' drops 6\% in AUC. Although it declines slightly, ``HiRAM w/o Rel Guidance'' does not perform well on top-n precision, especially in One setting.
Meanwhile, top-n precision of ``HiRAM w/o CFTE'' drops by nearly 10.5\%. 
To prove the superiority of pairwise types, the AUC score of ``HiRAM w/ Type Concat'' decreases by 6\% and nearly 5.6\% of top-n precision. 
Due to BERT's strong semantic learning ability, the AUC score of ``HiRAM w/ BERT-base'' outperforms HiRAM by 0.045 while its top-n precision has dropped by nearly 6\%, which indicates context-free pairwise types can increase accuracy and type-sentence alignment can enhance the confidence of prediction. 

\subsection{Performance on Long-Tail Relations} \label{sec:long-tail}
Since former baselines are mainly trained on NYT-570K, we reproduce CoRA on NYT-520K for fair comparison.
HiRAM achieves a new state-of-the-art result in Hits@K with 20\% superiority.
Removing hierarchy module in \S \ref{sec:hier}, the performance of ``HiRAM w/o Hierarchy'' decreases by nearly 30\% on Hits@10 but is better than baselines in other settings, verifying the importance of hierarchical model for long-tail relations. 
The huge decline of ``HiRAM w/o Rel Guidance'' verifies the necessity of relation guidance. 
The result of ``HiRAM w/ BERT-base'' is the worst in this ablation study for long-tail relations due to its dependence on sufficient training data with diverse semantics. This verifies that our specific embedding design is quite effective.

\subsection{Case Study and Error Analysis}
Firstly, we conduct a case study to qualitatively analyze the effect of our model in \S \ref{sec:hier} The case study of two samples are shown in Table \ref{tab:case_study} and the type-sentence alignment distribution is shown in Figure \ref{fig:case_study}.
Secondly, we investigate the possible reasons for the misclassifications of HiRAM.

\paragraph{Distribution of Type-Sentence Alignment.} 
For the first case, despite the failure in expressing the long-tail relation ``\textsc{/people/person/religion}'', the selected pairwise types are sufficient to predict this relation. 
As the top row of Figure \ref{fig:case_study} shows, \textsl{people.person} with \textsl{BLANK} helps to identify the character of subject entity, and \textsl{religion.religion} with high alignment score can provide direct attributes. 
For the second case, the semantics is implicitly related to its long-tail relation ``\textsc{/business/company/founder}''. The most proper pairwise types are selected with the hierarchical relation guidance, like (\textsl{organizer.organizer}, \textsl{organizer.founder}).

\paragraph{Error Analysis.} To analyse the implicit reasons for wrong predictions, we have manually checked several randomly-sampled error test examples. 
1) Most of error cases are annotated as \textsc{/people/Person/Place\_Of\_Birth} because the semantics and the relation may be completely irrelevant and the types of entities are hard to maintain people's birth place.
2) The global query in Eq.(\ref{eq:global_query}) could be invalid when the entity has too many characters. Mean pooling might not be the most suitable way to replace entity itself.

\section{Related Work}


\paragraph{Wrong Labeling Problem.}
Many works \citep{liu2016learning,ji2017distant,ye2019distant,li2020self} propose various extensions of vanilla selective attention \citep{lin2016neural}.
\citet{ye2019distant} combine intra-/inter-bag level selective attention for DSRE.
For one-sentence bags, \citet{li2020self} design the entity-aware embedding in a context-free manner with a gate mechanism. 

\paragraph{Long-tail Relations.} Knowledge transfer via hierarchical relations is effective.  \citet{han2018hierarchical} design relation-to-sentence attention in hierarchies, and \citet{li2020improving} modify it to sentence-to-relation attention. 
Many works \citep{vashishth2018reside,hu2019improving,chu2020insrl} resort to extra knowledge, i.e., entity description and entity types. 
Entity description \citep{hu2019improving, chu2020insrl} mainly stems from the Wikipedia page, which contains factual statements of the relation with other entities. 
Such oracle knowledge can boost DSRE performance but is impractical.

\section{Conclusion}
In this work, we propose a new model, HiRAM, to alleviate wrong labeling and long-tail problems in DSRE. 
For the wrong labeling problem, we propose a context-free type-enriched word embedding to enrich each word with prior knowledge and a context-related type-sentence alignment module to complement sentences with semantics-fitted pairwise types.
For the long-tail problem, we extend the base alignment into the hierarchy to utilize the multi-granular entity types.
The experiments with extensive analyses show the superiority of our HiRAM.


\bibliography{reference}

\end{document}